\pdfoutput=1

\documentclass[a4paper,conference]{IEEEtran}
\IEEEoverridecommandlockouts
\usepackage{times}
\usepackage{epsfig}
\usepackage{graphicx}
\usepackage{overpic}
\usepackage{booktabs}
\usepackage{multirow}
\usepackage{multicol}
\usepackage[noend]{algpseudocode}
\usepackage{algorithmicx,algorithm}
\usepackage{amsmath}
\usepackage{amssymb}
\usepackage{color}
\usepackage{balance}

\newcommand{\figref}[1]{Fig.~\ref{#1}}
\newcommand{\tabref}[1]{Tab.~\ref{#1}}
\newcommand{\equref}[1]{Equ.~(\ref{#1})}

\usepackage{bm}
\newcommand{\varMatrix}[1]{{\bm{#1}}}
\newcommand{\varVector}[1]{{\bm{#1}}}
\newcommand{\varScalar}[1]{#1}

\newcommand{\Transpose}[1]{{\rm{#1}}}
\newcommand{\Numdomain}[1]{{\mathbb{#1}}}

\usepackage[pagebackref=true,breaklinks=true,colorlinks,bookmarks=false]{hyperref}

\ifCLASSINFOpdf
\else
\fi
\begin{document}
%
\title{Seeking Salient Facial Regions for Cross-Database Micro-Expression Recognition}

\author{\IEEEauthorblockN{Xingxun Jiang, Yuan Zong\textsuperscript{\#}, Wenming Zheng\textsuperscript{\#}, Jiateng Liu, Mengting Wei}
\thanks{\# indicates the corresponding authors}
\IEEEauthorblockA{Key Laboratory of Child Development and Learning Science of Ministry of Education, \\
School of Biological Science and Medical Engineering, Southeast University, Nanjing 210096, China \\
\{jiangxingxun, xhzongyuan, wenming\_zheng, jiateng\_liu, weimengting\}@seu.edu.cn}
}


%


\maketitle

\begin{abstract}
Cross-Database Micro-Expression Recognition (CDMER) aims to develop the Micro-Expression Recognition (MER) methods with strong domain adaptability, i.e., the ability to recognize the Micro-Expressions (MEs) of different subjects captured by different imaging devices in different scenes.
The development of CDMER is faced with two key problems: 1) the severe feature distribution gap between the source and target databases; 2) the feature representation bottleneck of ME such local and subtle facial expressions.
To solve these problems, this paper proposes a novel Transfer Group Sparse Regression method, namely TGSR, which aims to 1) optimize the measurement and better alleviate the difference between the source and target databases, and 2) highlight the valid facial regions to enhance extracted features, by the operation of selecting the group features from the raw face feature, where each region is associated with a group of raw face feature, i.e., the salient facial region selection.
Compared with previous transfer group sparse methods, our proposed TGSR has the ability to select the salient facial regions, which is effective in alleviating aforementioned problems for better performance and reducing the computational cost at the same time.
We use two public ME databases, i.e., CASME II and SMIC, to evaluate our proposed TGSR method.
Experimental results show that our proposed TGSR learns the discriminative and explicable regions, and outperforms most state-of-the-art subspace-learning-based domain-adaptive methods for CDMER.
\end{abstract}


%
\IEEEpeerreviewmaketitle

\section{Introduction}
%
Micro-Expression (ME) is a low amplitude and short duration facial expression which may reflect subjects' genuine emotions\cite{ekman1969nonverbal,ekman2009telling}.
It is indispensable in many fields, such as criminal investigations\cite{yan2013fast}, clinical diagnosis\cite{frank2009see}, human-computer interaction\cite{jiang2020dfew,li2019bi}, etc.
%

Due to the huge potential value of MEs, many efforts have been made to design an automatic Micro-Expression Recognition (MER) system over the last few decades\cite{wang2014lbp,wang2015micro,liu2015main,lu2016micro,zong2018learning}.
They developed many subspace-learning-based methods\cite{zhao2007dynamic,wang2014lbp,xu2017microexpression} and deep learning methods\cite{wei2022novel,xia2021motion}, and promoted the rapid development of automated MER technology.
However, most existing methods are evaluated on a single database, which may sharply drop the performance when applied in domains different from the training database, such as imaging devices, subjects, scenes, etc.

%
To learn a domain-robust MER model, researchers have turned their interests to the domain-adaptive MER method recently.
A new challenging topic has thus emerged, i.e., Cross-Database Micro-Expression Recognition (CDMER).
It mimics the domain variation problem and evaluates the method's adaptive ability by the operation of training the model in one micro-expression database, i.e., the source database, and testing in the other one, i.e., the target database.
CDMER\cite{zong2019toward,li2019unsupervised,chen2019target} is faced with two problems: 1) the severe feature distribution gap between the source and target databases, and 2) the feature representation bottleneck of ME such a subtle and local facial expressions.
Past research has proposed many subspace-learning-based methods to yield a similar and effective feature to bridge this gap between the source and target domains and effectively advance the MER model's adaptive ability.
However, existing CDMER models' performance is still far from satisfactory.

Salient facial region is an approach to enhancing the few but discriminative regions and suppressing the many but noisy regions, aiming to improve performance and reduce computational cost simultaneously.
It has been widely validated that the salient region selection approach benefits emotion recognition performance.
Inspired by this, we introduce a learnable binary sparse regression matrix shared between the source and target databases, and propose a novel Transfer Group Sparse Regression method (TGSR) to cope with the CDMER problem with the assistance of salient facial region selection technology.
Our proposed TGSR contains three terms: a regression term with the learnable matrix for bridging micro-expression features and labels, a joint feature distribution regularization term for measuring and alleviating the difference between source and target databases, and a regression matrix sparse term to promote our proposed TGSR learns the few but discriminative region feature.
Especially, the salient facial region selection in our proposed TGSR is achieved by the operation of selecting the group features from the raw features, where each region is associated with a group of raw face features.
And the facial region selection is aimed to seek the discriminative regions for 1) optimizing the measurement and better alleviating the difference between the source and target databases, and 2) highlighting the valid facial regions to enhance extracted features, which will significantly improve the performance of CDMER model.
In addition, facial region selection can also reduce the computational cost when pursuing better CDMER performance.
We evaluate our method on CASME II\cite{yan2014casme} and SMIC\cite{li2013spontaneous} databases.
Experimental results and corresponding visualization show that our proposed TGSR can seek the salient and explicable facial regions to alleviate the aforementioned problems effectively and outperform most state-of-the-art subspace-learning-based domain-adaptive methods for CDMER.

\section{Method}
\subsection{The Generation of Micro-Expression Features}
\begin{figure}[htbp]
   \centering
   \includegraphics[width=\columnwidth]{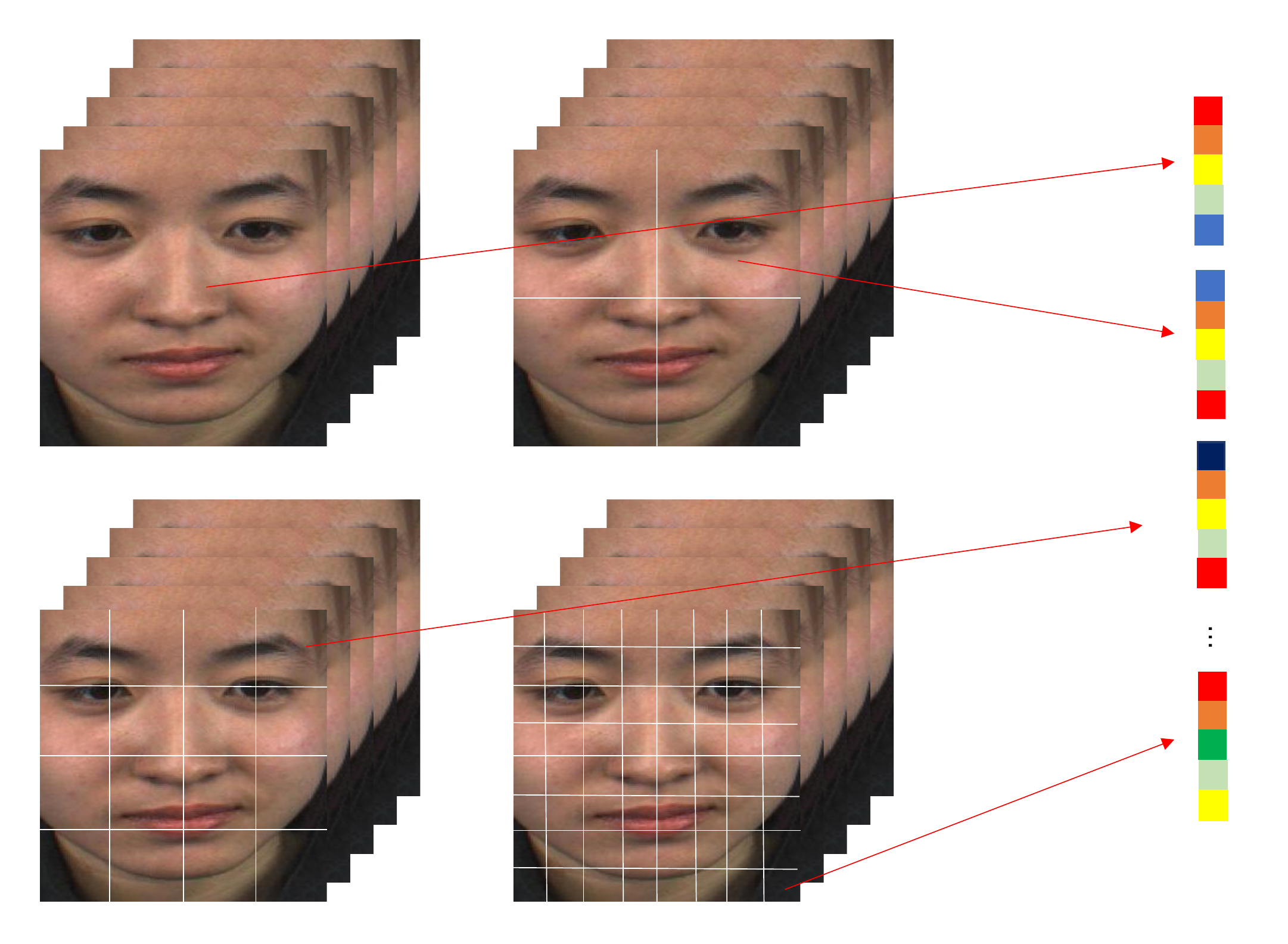}
   \caption{The grid-based multi-scale spatial division scheme for extracting micro-expression features.}
   \label{fig:multiGridFea}
\end{figure}

%
Extracting facial features is the first step for CDMER.
As \figref{fig:multiGridFea} shown, we firstly use the grid-based multi-scale spatial division scheme\cite{zhang2020cross} to divide the cropped ME sequence into four scales in total of $K$ regions, i.e., $K$ spatial local sequences.
Then we extracted $d$-dimensional feature $\varVector{x}_k$ of $K$ facial region, $k\in[1,K]$, and obtain sample's multi-scale hierarchical feature $\varVector{x}^{\nu}=\left[\varVector{x}_{1}^{\Transpose{T}},\cdots,\varVector{x}_{K}^\Transpose{T}\right]^\Transpose{T} \in \Numdomain{R}^{Kd}$ by concatenating region features one by one.
Suppose that we have $N_s$ source and $N_t$ target micro-expression samples, the feature matrix of the source and target databases can be denoted as $\varMatrix{X}^s=\left[ {\varMatrix{X}_1^s}^\Transpose{T}, \cdots, {\varMatrix{X}_K^s}^\Transpose{T} \right]^\Transpose{T} \in \Numdomain{R}^{\varScalar{K}\varScalar{d}\times\varScalar{N_s}}$ and $\varMatrix{X}^t=\left[ {\varMatrix{X}_1^t}^\Transpose{T},\cdots,{\varMatrix{X}_K^t}^\Transpose{T} \right]^\Transpose{T} \in \Numdomain{R}^{\varScalar{K}\varScalar{d}\times\varScalar{N_t}}$, respectively.
Here, each column of $\varMatrix{X}^s$ and $\varMatrix{X}^t$ is a feature vector like $\varVector{x}^{\nu}$, they respectively denote the feature of single micro-expression sample from the corresponding databases.
$\varMatrix{X}_i^s\in\Numdomain{R}^{d\times\varScalar{N_s}}$ and $\varMatrix{X}_i^t \in \Numdomain{R}^{d\times\varScalar{N_t}}$ respectively denote the group feature corresponding to the $i$-th facial region from the source and target databases.
The labels of source micro-expression database is denoted by $\varMatrix{L}^s=[\varVector{l}^s_1,\cdots,\varVector{l}^s_{N_s}]\in\Numdomain{R}^{C\times\varScalar{N_s}}$, where $C$ is the total category number and the $j$-th column of $\varMatrix{L}^s$ denotes the label vector of $j$-th source micro-expression sample.
The label vector of $j$-th sample $\varVector{l}^s_j=[l^s_{j,1},\cdots,l^s_{j,C}]^\Transpose{T}$ is a one-hot vector in which only one element $l^s_{j,c}$ equals one and the others are zero.
It indicates that $j$-th sample from the source database belongs to $c$-th micro-expression category.

\subsection{Proposed Method}
%
%
The basic idea of our proposed Transfer Group Sparse Regression method (TGSR) is improve the CDMER performance by sailent facial region selection approach, i.e., selecting the group features from the raw face feature, where each region is associated with a group of raw face feature.
Our proposed TGSR contains three terms: 1) a regression term with the learnable regression matrix for bridging micro-expression features and labels, 2) a joint feature distribution regularization term for measuring and alleviating the difference between source and target databases, and 3) a regression matrix sparse term to promote our proposed TGSR learns the few but discriminative region feature, which can be denoted as \equref{equ:LinearRegressionZong},
%
%
\begin{equation}\label{equ:LinearRegressionZong}
\min \limits_{\varMatrix{C}_i} \left\| \varMatrix{L}^s-\sum \limits_{i=1}^K \varMatrix{C}_i^\Transpose{T}\varMatrix{X}_i^s \right\|_F^2 + \xi f_1(\varMatrix{C}_i) + \lambda f_2(\varMatrix{C}_i),
\end{equation}
where $\varMatrix{C}=[\varMatrix{C}^{\Transpose{T}}_{1}, ..., \varMatrix{C}^{\Transpose{T}}_{K}]^{\Transpose{T}} \in \Numdomain{R}^{{Kd}\times\varScalar{C}}$ is such a domain-invariant regression matrix, the sub-matrix $\varMatrix{C}_i$ of $\varMatrix{C}$ is to construct the relation between the group feature of $i$-th facial region and corresponding sample labels, $f_1(\varMatrix{C}_i)$ is the joint feature distribution regularization term, $f_2(\varMatrix{C}_i)$ is the regression matrix sparse term, and $\varScalar{\xi}$ and $\varScalar{\lambda}$ the weighting hyper-parameters of $f_1(\varMatrix{C}_i)$ and $f_2(\varMatrix{C}_i)$.

%
%
By minimizing $f_1(\varMatrix{C}_i)$ together with the regression term, we can alleviate the database difference.
We use the maximum mean discrepancy (MMD) to serve as this regularization term, which can be expressed as \equref{equ:MMD1},
\begin{equation}\label{equ:MMD1}
\begin{split}
& MMD\left(\varMatrix{X}^s,\varMatrix{X}^t\right)=\\
& \left\|\frac{1}{N_s}\sum_{i=1}^{K}{\mit\Phi}\left(\varMatrix{X}_i^s\right)\varVector{1}_{s}-\frac{1}{N_t}\sum_{i=1}^K{\mit\Phi}\left(\varMatrix{X}^t\right)\varVector{1}_{t}\right\|_\mathcal{H},\\
\end{split}
\end{equation}
where $\mit{\Phi}(\cdot)$ is a kernel mapping operator projecting micro-expression features from the original space to an infinite one, $\varVector{1}_s \in \Numdomain{R}^{N_s}$ and $\varVector{1}_t \in \Numdomain{R}^{N_t}$ are the vectors filled with scalar value one which used to convert the source and target features into scalar values respectively. 
However, the kernel mapping operator is unsolvable, so we further modify the MMD into \equref{equ:MMD2} to serve as $f_1(\varMatrix{C}_i)$,
\begin{equation}\label{equ:MMD2}
f_1\left(\varMatrix{C}_i\right)=
\left\|\frac{1}{N_s}\sum_{i=1}^K\varMatrix{C}_i^\Transpose{T}\varMatrix{X}_i^s\varVector{1}_s-\frac{1}{N_t}\sum_{i=1}^K\varMatrix{C}_i^\Transpose{T}\varMatrix{X}_i^t\varVector{1}_t\right\|_F^2.
\end{equation}
By relaxing the difference measurement involving kernel-mapped feature into the difference in label space, \equref{equ:MMD1} becomes solvable.
$f_2(\varMatrix{C}_i)$ is defined as \equref{equ:SparseTerm}.
Our proposed TGSR is promoted to select the few but discriminative regions when minimizing $f_2(\varMatrix{C}_i)$ together with the regression matrix.
\begin{equation}\label{equ:SparseTerm}
f_2\left(\varMatrix{C}_i\right)=\lambda \sum \limits_{i=1}^K \left\|\varMatrix{C}_i\right\|_F.
\end{equation}

By substituting \equref{equ:MMD2} and \equref{equ:SparseTerm} into \equref{equ:LinearRegressionZong}, we can rewrite the objective function of \equref{equ:LinearRegressionZong} into \equref{equ:final1Ori},
\begin{equation}\label{equ:final1Ori}
\begin{split}
& \min_{\varMatrix{C}_i} 
\left\|\varMatrix{L}^s -\sum_{i=1}^K\varMatrix{C}_i^\Transpose{T}\varMatrix{X}_i^s\right\|_F^2 + \lambda\sum_{i=1}^K \left\| \varMatrix{C}_i \right\|_F\\
& +\xi \left\|\frac{1}{N_s}\sum_{i=1}^K\varMatrix{C}_i^\Transpose{T}\varMatrix{X}_i^s\varVector{1}_s-\frac{1}{N_t}\sum_{i=1}^K\varMatrix{C}_i^\Transpose{T}\varMatrix{X}_i^t\varVector{1}_t\right\|_F^2.\\
\end{split}
\end{equation}
%



\subsection{Optimization}
We can use the Alternative Direction Method (ADM)\cite{qin2012structured} and Inexact Augmented Lagrangian Multiplier (IALM)\cite{lin2010augmented} to solve \equref{equ:final1Ori} involving facial region selection technology, i.e., selecting the group features of the raw face features, under presetting salient region number.

We firstly rewrite \equref{equ:final1Ori} into \equref{equ:final5MatrixForm},
\begin{equation}\label{equ:final5MatrixForm}
\min_{\varMatrix{C}_i}
\left\| \tilde{\varMatrix{L}}-\sum_{i=1}^{K} \varMatrix{C}_i^\Transpose{T} \tilde{\varMatrix{X}}_i \right\|_F^2
+ \varScalar{\lambda} \sum \limits_{i=1}^K \left\| \varMatrix{C}_i \right\|_F,
\end{equation}
where $\tilde{\varMatrix{L}} = \left[\varMatrix{L}^s,\varVector{0}\right]$, $\varVector{0} \in \Numdomain{R}^{C\times1}$, $\tilde{\varMatrix{X}}_i = \left[\varMatrix{X}_i^s, \sqrt{\varScalar{\xi}} (\frac{1}{N_s}\varMatrix{X}_i^s\varVector{1}_s-\frac{1}{N_t}\varMatrix{X}_i^t\varVector{1}_t) \right]$.
Then we introduce a new variable $\varMatrix{D}=[\varMatrix{D}_1^\Transpose{T}, \cdots, \varMatrix{D}_K^\Transpose{T}]^\Transpose{T}$ equals to variable $\varMatrix{C}=[\varMatrix{C}_i^\Transpose{T},\cdots,\varMatrix{C}_K^\Transpose{T}]^\Transpose{T}$, and convert the optimization of \equref{equ:final5MatrixForm} into a constrained one as \equref{equ:final6stlimit},
%
\begin{equation}\label{equ:final6stlimit}
\begin{array}{l}
\min \limits_{\varMatrix{C},\varMatrix{D}}
\left\| \tilde{\varMatrix{L}} - \sum \limits_{i=1}^{K} \varMatrix{D}_i^\Transpose{T} \tilde{\varMatrix{X}}_i \right\|_F^2
+ \varScalar{\lambda} \sum\limits_{i=1}^{K} \left\| \varMatrix{C}_i \right\|_F,\\
\operatorname{ s.t. }\varMatrix{D}_i=\varMatrix{C}_i.\\
\end{array}
\end{equation}
Subsequently, we can obtain the corresponding augmented Lagrange function as \equref{equ:largarian} shown,
\begin{equation}\label{equ:largarian}
\begin{split}
{\mit{\Gamma}}\left(\varMatrix{C}_i,\varMatrix{D}_i,\varMatrix{P}_i,\varScalar{\mu}\right)=
\left\| \tilde{\varMatrix{L}}-\sum \limits_{i=1}^{K}\varMatrix{D}_i^\Transpose{T}\tilde{\varMatrix{X}}_i \right\|_F^2
+\varScalar{\lambda} \sum \limits_{i=1}^{K} \left\|\varMatrix{C}_i\right\|_F \\
+\sum \limits_{i=1}^{K} \operatorname{tr} \left[\varMatrix{P}_i^\Transpose{T} \left(\varMatrix{C}_i-\varMatrix{D}_i\right) \right]
+\frac{\varScalar{\mu}}{2} \sum \limits_{i=1}^{K} \left\| \varMatrix{C}_i-\varMatrix{D}_i \right\|_F^2,
\end{split}
\end{equation}
\noindent where $\varMatrix{P}_i \in \Numdomain{R}^{d\times{C}}$ denotes the Lagrangian multiplier matrix corresponding to the $i$-th facial region, and $\mu$ is the weighting hyper-parameter.

We can obtain the optimal solution $\hat{\varMatrix{C}_i}$ of $\varMatrix{C}_i$ when minimizing the Lagrange function of \equref{equ:largarian} by iteratively update $\varMatrix{C}_i$ and $\varMatrix{D}_i$.
%
Specifically, we need to repeat the following four steps until convergence, and \textit{Algorithm 1} show more details:

1) Fix $\varMatrix{C}$, $\varMatrix{P}$, $\mu$ and update $\varMatrix{D}$: 

In this step, the optimization problem with respect to the sub-matrix $\varMatrix{D}_i$ of $\varMatrix{D}$ can be written as \equref{equ:lan1},
\begin{equation}\label{equ:lan1}
\begin{split}
\min\limits_{\varMatrix{D}} \left\| \tilde{\varMatrix{L}} - \varMatrix{D}^\Transpose{T} \tilde{\varMatrix{X}} \right\|_F^2 + \operatorname{tr}\left[\varMatrix{P}^\Transpose{T} \left(\varMatrix{C}-\varMatrix{D}\right) \right] +\frac{\varScalar{\mu}}{2}\left\| \varMatrix{C}-\varMatrix{D} \right\|_F^2, 
\end{split}
\end{equation}
\noindent where $\varMatrix{P}^\Transpose{T}=[\varMatrix{P}_1^\Transpose{T},\cdots,\varMatrix{P}_K^\Transpose{T}]$, $\varMatrix{P} \in \Numdomain{R}^{Kd\times C}$, $\varMatrix{P}_j \in \Numdomain{R}^{d\times C}$. The closed-form solution of \equref{equ:lan1} as \equref{equ:calMatrixD} shows.

2) Fix $\varMatrix{D}$, $\varMatrix{P}$, $\mu$ update $\varMatrix{C}$:

In this step, the optimization problem with respect to the sub-matrix $\varMatrix{C}_i$ of $\varMatrix{C}$ can be written as \equref{equ:lan2},
\begin{equation}\label{equ:lan2}
\begin{split}
\min\limits_{\varMatrix{C}_i} \varScalar{\lambda} \sum\limits_{i=1}^K\left\|\varMatrix{C}_i\right\|_F +\sum\limits_{i=1}^K \operatorname{tr}\left[\varMatrix{P}_i^\Transpose{T} \left(\varMatrix{C}_i-\varMatrix{D}_i\right) \right] \\
+\frac{\varScalar{\mu}}{2} \sum\limits_{i=1}^K \left\| \varMatrix{C}_i-\varMatrix{D}_i \right\|_F^2.
\end{split}
\end{equation}
We can convert \equref{equ:lan2} into \equref{equ:lan3}, and obtain the optimal $\varMatrix{C}$ using \equref{equ:calMatrixC}.
\begin{equation}\label{equ:lan3}
\min\limits_{\varMatrix{C}_i} \sum\limits_{i=1}^K ( \frac{\lambda}{\mu} \left\| \varMatrix{C}_i \right\|_F + \frac{1}{2} \left\| \varMatrix{C}_i-( \varMatrix{D}_i - \frac{\varMatrix{P}_i}{\mu} ) \right\|_F^2 )
\end{equation}

3) Update $\varMatrix{P}$ and $\mu$.

4) Check the convergence of $\left\|\varMatrix{C}-\varMatrix{D}\right\|_{\infty}<\varepsilon$.

\begin{algorithm}\label{alg:optimization}
   \caption{The Algorithm for solving the optimal regression matrix $\varMatrix{C}$ in our proposed TGSR method.}

   \hspace*{0.02in}{
   \textbf{Input:} Data matrix $\tilde{\varMatrix{L}}$ and $\tilde{\varMatrix{X}}=[\tilde{\varMatrix{X}}_1^\Transpose{T},\cdots,\tilde{\varMatrix{X}}_K^\Transpose{T}]^\Transpose{T}$, 
   the salient facial region number $\kappa$, 
   the scalar parameter $\varScalar{\rho}$,
   $\varScalar{\mu_{max}}$.
   }

   \begin{itemize}
      \item Initializing the regression matrix $\varMatrix{C}=[\varMatrix{C}_1^\Transpose{T},\cdots,\varMatrix{C}_K^\Transpose{T}]^\Transpose{T}$
      \item Initializing the Lagrangian multiplier matrix $\varMatrix{P}=[\varMatrix{P}_1^\Transpose{T},\cdots,\varMatrix{P}_K^\Transpose{T}]^\Transpose{T}$ and the weighting coefficient $\varScalar{\mu}$.
   \end{itemize}

   \hspace*{0.02in}{\bf Repeating steps 1) to 4) until convergence.}
   \begin{algorithmic}[1]
   \State Fix $\varMatrix{C},\varMatrix{P},\varScalar{\mu}$ and update $\varMatrix{D}$:
   \begin{equation}\label{equ:calMatrixD}
   \varMatrix{D}=\left(\varScalar{\mu}\varMatrix{I}_{Kd} + 2\tilde{\varMatrix{X}}\tilde{\varMatrix{X}}^\Transpose{T}\right)^{-1}\left(2\tilde{\varMatrix{X}}\tilde{\varMatrix{L}}^{\Transpose{T}}+\varMatrix{P}+\varScalar{\mu}\varMatrix{C}\right),
   \end{equation}
   
   \noindent where $\varMatrix{I}_{Kd} \in \Numdomain{R}^{Kd\times Kd}$ is the identity matrix.
   \State Fix $\varMatrix{D}, \varMatrix{P}, \varScalar{\mu}$ and update $\varMatrix{C}$:

   Calculate $\varScalar{d}_i=\left\| \varMatrix{D}_i - \cfrac{\varMatrix{P}_i}{\varScalar{\mu}}\right\|_F$, 
   and sort the value of 
   $\varScalar{d}_i$, such that $\varScalar{d}_{i_{1}}\geq\varScalar{d}_{i_{2}}\geq\cdots\geq\varScalar{d}_{i_{K}}$, 
   Let $\varScalar{\lambda}=\varScalar{\mu}\varScalar{d}_{i_{\kappa+1}}$, 
   then update $\varMatrix{C}$ according to

   \begin{equation}\label{equ:calMatrixC}
   \varMatrix{C}_i=
   \begin{cases}
   \dfrac{\varScalar{d}_i-\frac{\varScalar{\lambda}}{\varScalar{\mu}}}{\varScalar{d}_i}(\varMatrix{D}_i-\dfrac{\varMatrix{P}_i}{\varScalar{\mu}}),&\frac{\varScalar{\lambda}}{\varScalar{\mu}}<\varScalar{d}_i,\\
   \varVector{0},&\frac{\varScalar{\lambda}}{\varScalar{\mu}}\geq\varScalar{d}_i.
   \end{cases}
   \end{equation}

   \State Update $\varMatrix{P}$ and $\mu$:

   $\varMatrix{P}=\varMatrix{P}+\varScalar{\mu}\left(\varMatrix{D}-\varMatrix{C}\right)$, $\varScalar{\mu}=\min{\left(\varScalar{\rho}\varScalar{\mu},\varScalar{\mu}_{max}\right)}$

   \State Check convergence:

   $\left\|\varMatrix{C}-\varMatrix{D}\right\|_{\infty}<\varepsilon$

   \end{algorithmic}

   \hspace*{0.02in}{
   \textbf{Output:} The solution $\hat{\varMatrix{C}}$ of regression matrix $\varMatrix{C}$.
   }

\end{algorithm}

\subsection{Application for CDMER}
Based on the labeled source and the unlabeled target databases, we can solve the optimal solution $\hat{\varMatrix{C}}$ of regression matrix $\varMatrix{C}$ using aforementioned optimization approach.
Then, we can extract the feature $\varVector{x}_i^{te} \in \Numdomain{R}^{Kd}$ of the micro-expression sample to be predicted and estimate the label vector $\varVector{l}^{te}$ by solving the optimization problem as \equref{equ:testEQU},

\begin{equation}\label{equ:testEQU}
\begin{array}{l}
\min \limits_{\varVector{l}^{te}} \left\| \varVector{l}^{te} - \sum \limits_{i=1}^{K} \hat{\varMatrix{C}}_i^\Transpose{T} \varVector{x}_i^{te} \right\|_F^2,\\
\operatorname{ s.t. } \quad \varVector{l}^{te}\geq0. \varVector{1}^\Transpose{T}\varVector{l}^{te}=1,
\end{array} 
\end{equation}

\noindent where $\hat{\varMatrix{C}}_i \in \Numdomain{R}^{d\times{C}}$ is the optimal solution of the regression matrix for the $i$-th facial spatial local region, and $\hat{\varMatrix{C}}^\Transpose{T}=\left[\hat{\varMatrix{C}}_1^\Transpose{T},\cdots,\hat{\varMatrix{C}}_K^\Transpose{T}\right]$, $\hat{\varMatrix{C}}^\Transpose{T} \in \Numdomain{R}^{C\times{Kd}}$, $\varVector{l}^{te} \in \Numdomain{R}^C$.
Then we can use $\hat{c} = \arg \max_j \left\{ \varVector{l}_j^{te} \right\}$ to assign this micro-expression sample to the largest entry index of the predicted label vector, i.e., micro-expression category $\hat{c}$.


\begin{table}[htbp]
   \centering
   \caption{The statistics of Selected CASME II and SMIC database.}
   \label{tab:datasetStat}
   \begin{tabular}{cccc}
   \toprule
   \multirow{2}{*}{Dataset}& \multicolumn{3}{c}{Category}\\
   \cline{2-4}
   &Positive & Negative & Surprise \\
   \midrule
   Selected CASME II & 32 & 73 & 25\\
   SMIC-HS&51&70&43\\
   SMIC-VIS&23&28&20\\
   SMIC-NIR&23&28&20\\
   \bottomrule
   \end{tabular}
\end{table}

\begin{table*}[htbp]
   \centering
   \caption{
   The results of TYPE-I CDMER experiments are based on any two subsets of SMIC, i.e., SMIC-HS, SMIC-VIS, and SMIC-NIR.
   The micro-expression category includes \textit{Negative}, \textit{Positive}, and \textit{Surprise}.
   The best results from each experiment are shown in bold.
   We use Macro F1-score (M-F1) and Accuracy (ACC) to evaluate methods.
   }
   
   \label{tab:exp1}

   \begin{tabular}{lcccccccccccccc}
    \toprule 
   \multirow{2}{*}{Method} & \multicolumn{2}{c}{Exp.1(H$\rightarrow$V)} & \multicolumn{2}{c}{Exp.2(V$\rightarrow$H)} & \multicolumn{2}{c}{Exp.3(H$\rightarrow$N)} & \multicolumn{2}{c}{Exp.4(N$\rightarrow$H)} & \multicolumn{2}{c}{Exp.5(V$\rightarrow$N)} & \multicolumn{2}{c}{Exp.6(N$\rightarrow$V)} & \multicolumn{2}{c}{Average}\\
   \cline{2-15}
   &M-F1&ACC&M-F1&ACC&M-F1&ACC&M-F1&ACC&M-F1&ACC&M-F1&ACC&M-F1&ACC\\
    \midrule
   SVM\cite{chang2011libsvm} & 0.8002 & 80.28  & 0.5421 & 54.27  & 0.5455 & 53.52 & 0.4878 & 54.88 & 0.6186 & 63.38 & 0.6078 & 63.38 & 0.6003 & 61.62\\
   IW-SVM\cite{hassan2013acoustic} & 0.8868 & 88.73 & 0.5852 & 58.54 & \textbf{0.7469} & \textbf{74.65} & 0.5427 & 54.27 & 0.6620 & 69.01 & 0.7228 & 73.24 & 0.6911 & 67.74\\
   TCA\cite{pan2010domain} & 0.8269 & 83.10 & 0.5477 & 54.88 & 0.5828 & 59.15 & 0.5443 & 57.32 & 0.5810 & 61.97 & 0.6598 & 67.61 & 0.6238 & 64.01\\
   GFK\cite{gong2012geodesic} & 0.8448 & 84.51 & 0.5957 & 59.15 & 0.6977 & 70.42 & 0.6197 & \textbf{62.80} & \textbf{0.7619} & \textbf{76.06} & 0.8142 & 81.69& \textbf{0.7223} & \textbf{72.44}\\
   SA\cite{fernando2013unsupervised}  & 0.8037 & 80.28 & 0.5955 & 59.15 & 0.7465 & \textbf{74.65} & 0.5644 & 56.10 & 0.7004 & 71.83 & 0.7394 & 74.65 & 0.6917 & 69.44\\
   STM\cite{chu2013selective} & 0.8253 & 83.10 & 0.5059 & 51.22 & 0.6628 & 66.20 & 0.5351 & 56.10 & 0.6427 & 67.61 & 0.6922 & 70.42 & 0.6440 & 65.78\\
   TKL\cite{long2014domain} & 0.7742 & 77.46 & 0.5738 & 57.32 & 0.7051 & 70.42 & 0.6116 & 62.20 & 0.7558 & 76.06 & 0.7580 & 76.06 & 0.6964 & 69.92\\
   TSRG\cite{zong2017learning}& 0.8869 & 88.73 & 0.5652 & 56.71 & 0.6484 & 64.79 & 0.5770 & 57.93 & 0.7056 & 70.42 & 0.8116 & 81.69 & 0.6991 & 70.05\\
   DRLS\cite{zong2018domain}& 0.8604 & 85.92 & 0.6120 & 60.98 & 0.6599 & 66.20 & 0.5599 & 55.49 & 0.6620 & 69.01 & 0.5771 & 61.97& 0.6552 & 66.60\\ 
   \midrule
   Ours& \textbf{0.9150} & \textbf{91.55} & \textbf{0.6226} & \textbf{62.20} & 0.5847 & 60.56 & \textbf{0.6272} & 61.59 & 0.6984 & 70.42 & \textbf{0.8403} & \textbf{84.51} & 0.7141 & 71.80\\   
    \bottomrule
   \end{tabular}
\end{table*}

\begin{table*}[htbp]
   \centering
   \caption{
   The results of TYPE-II CDMER experiments are based on Selected CASME II database and any one subset of SMIC databases, i.e., one of SMIC-HS, SMIC-VIS, and SMIC-NIR.
   The micro-expression category includes \textit{Negative}, \textit{Positive}, and \textit{Surprise}.
   The best results from each experiment are shown in bold.
   We use Macro F1-score (M-F1) and Accuracy (ACC) to evaluate methods.
   }
   \label{tab:exp2}

   \begin{tabular}{lcccccccccccccccc}
   \toprule
   \multirow{2}{*}{Method} & \multicolumn{2}{c}{Exp.7(C$\rightarrow$H)}& \multicolumn{2}{c}{Exp.8(H$\rightarrow$C)}& \multicolumn{2}{c}{Exp.9(C$\rightarrow$V)}& \multicolumn{2}{c}{Exp.10(V$\rightarrow$C)} & \multicolumn{2}{c}{Exp.11(C$\rightarrow$N)} & \multicolumn{2}{c}{Exp.12(N$\rightarrow$C)}&\multicolumn{2}{c}{Average}\\
   \cline{2-15}
   &M-F1&ACC&M-F1&ACC&M-F1&ACC&M-F1&ACC&M-F1&ACC&M-F1&ACC&M-F1&ACC\\
   \midrule
   SVM\cite{chang2011libsvm} & 0.3697 & 45.12 & 0.3245 & 48.46 & 0.4701 & 50.70 & 0.5367 & 53.08 & 0.5295 & 52.11 & 0.2368 & 23.85& 0.4112 & 45.55\\
   IW-SVM\cite{hassan2013acoustic} & 0.3541 & 41.46 & 0.5829 & 62.31 & 0.5778 & 59.15 & 0.5537 & 54.62 & 0.5117 & 50.70 & 0.3456 & 36.15 & 0.4876 & 50.73\\
   TCA\cite{pan2010domain} & 0.4637 & 46.34 & 0.4870 & 53.08 & \textbf{0.6834} & \textbf{69.01} & 0.5789 & 59.23 & 0.4992 & 50.70 & 0.3937 & 42.31 & 0.5177 & 53.45\\
   GFK\cite{gong2012geodesic} & 0.4126 & 46.95 & 0.4776 & 50.77 & 0.6361 & 66.20 & 0.6056 & 61.50 & 0.5180 & 53.52 & 0.4469 & 46.92 & 0.5161 & 54.31\\
   SA\cite{fernando2013unsupervised}  & 0.4302 & 47.56 & 0.5447 & 62.31 & 0.5939 & 59.15 & 0.5243 & 51.54 & 0.4738 & 47.89 & 0.3592 & 36.92 & 0.4877 & 50.90\\
   STM\cite{chu2013selective} & 0.3640 & 43.90 & \textbf{0.6115} & \textbf{63.85} & 0.4051 & 52.11 & 0.2715 & 30.00 & 0.3523 & 42.25 & 0.3850 & 41.54 & 0.3982 & 45.61\\
   TKL\cite{long2014domain} & 0.4582 & 46.95 & 0.4661 & 54.62 & 0.6042 & 60.56 & 0.5378 & 53.08 & 0.5392 & 54.93 & 0.4248 & 43.85 & 0.5051 & 52.33\\
   TSRG\cite{zong2017learning}& \textbf{0.5042} & 51.83 & 0.5171 & 60.77 & 0.5935 & 59.15 & 0.6208 & 63.08 & 0.5624 & 56.34 & 0.4105 & 46.15 & 0.5348 & 56.22\\
   DRLS\cite{zong2018domain}& 0.4924 &\textbf{53.05} & 0.5267 & 59.23 & 0.5757 & 57.75 & 0.5942 & 60.00 & 0.4885 & 49.83 & 0.3838 & 42.37 & 0.5102 & 53.71\\
   \midrule
   Ours& 0.5001 & 51.83 & 0.5061 & 56.92 & 0.5906 & 59.15 & \textbf{0.6403} & \textbf{63.85} & \textbf{0.5697} & \textbf{57.75} & \textbf{0.4474} & \textbf{48.46} & \textbf{0.5424} & \textbf{56.33}\\
   \bottomrule
   \end{tabular}
\end{table*}

\section{Experiment}
\subsection{Experiment Setup}
\subsubsection{Database}
We evaluated our method on Selected CASME II and SMIC databases. 
CASME II\cite{yan2014casme} contains 255 micro-expression samples from 26 subjects with seven category micro-expressions, i.e., \textit{Disgust}, \textit{Fear}, \textit{Happiness}, \textit{Others}, \textit{Repression}, \textit{Sadness}, and \textit{Surprise}. 
We selected the samples of \textit{Disgust}, \textit{Happiness}, \textit{Repression}, and \textit{Surprise} to be the Selected CASME II.
SMIC\cite{li2013spontaneous} records 306 micro-expression samples from 16 subjects in three modalities with three category micro-expressions, i.e., \textit{Positive}, \textit{Negative}, and \textit{Surprise}.
The SMIC-HS subset contains 164 micro-expression samples captured by a high-speed camera at 100 frames/s. 
The SMIC-VIS subset contains 71 micro-expression samples captured by a general visual camera at 25 frames/s.
The SMIC-NIR subset contains 71 micro-expression samples captured by a near-infrared camera.
In order to make the Selected CASME II and SMIC databases share the same label categories, we converted the labels of Selected CASME II: relabelled the label \textit{Happiness} into \textit{Positive}; relabelled the labels \textit{Disgust} and \textit{Repression} into \textit{Negative}; maintained the label \textit{Surprise} with \textit{Surprise}.
\tabref{tab:datasetStat} summarize the essential information.

\subsubsection{Protocol}
The cross-database protocol is designed to develop models with promising domain adaption performance operated by training the model in the Source database (S) and testing in the Target database (T), which is denoted as S$\rightarrow$T.
Following \cite{zhang2020cross}, we employed two types of unsupervised CDMER experiments: TYPE-I is implemented between every two subsets of SMIC, and TYPE-II is implemented between Selected CASME II and any subset of SMIC.
We denote SMIC-HS, SMIC-VIS, and SMIC-NIR as H, V, N, and CASME II as C for short.
Specially, TYPE-I experiment includes six experiments: H$\rightarrow$V, V$\rightarrow$H, H$\rightarrow$N, N$\rightarrow$H, V$\rightarrow$N, N$\rightarrow$V, TYPE-II experiment consists of another six experiments: C$\rightarrow$H, H$\rightarrow$C, C$\rightarrow$V, V$\rightarrow$C, C$\rightarrow$N, N$\rightarrow$C.

\subsubsection{Evaluation Metrics} 
We employed macro F1-score (M-F1) and accuracy (ACC) to evaluate our method.
Macro F1-score is calculated by $M-F1=\frac{1}{C} \sum_{c=1}^C \frac{2p_{c}r_{c}}{p_{c}+r_{c}}$, where $p_{c}$ and $r_{c}$ are the precision and recall of the $c$-th category micro-expression, and $C$ is the category number.
M-F1 is appropriate because the unbalanced sample problem widely existed in the CDMER.

\subsubsection{Data pre-processing}
We firstly cropped the whole face of each ME sequence using the bounding box from the first frame.
Then we employed the Temporal Interpolation Model (TIM)\cite{zhou2011towards, zhou2013compact} to convert the ME sequence into fixed 16 frames in temporal.
And resized each frame into $112\times112$ pixels in spatial.

\subsubsection{Feature extraction}
For each ME sequence, we used a grid-based multi-scale spatial division scheme to divide the whole face into four scales of $1\times1$, $2\times2$, $4\times4$, $8\times8$, a total of $K=85$ local face sequences, i.e., facial regions.
Then extracted and concatenated the corresponding LBP-TOP features\cite{zhao2007dynamic} of these facial regions to serve as the micro-expression representation.
Here, the neighboring radius of LBP-TOP and the number of neighboring points are set to $R=3$ and $P=8$.

\subsubsection{Training setting}
Two hyper-parameters are involved in solving our proposed TGSR, i.e., the salient facial region number $\kappa$ and the weighting hyper-parameter $\xi$ of the MMD term.
%
%
Following the work of \cite{zong2017learning,zhang2020cross}, we used a grid-based searching strategy to search the optimal hyper-parameters of our proposed TGSR for achieving the best M-F1 performance.
We reported both M-F1 and ACC metrics under the optimal setting.
Specially, we searched the hyper-parameter $\kappa$ from a preset parameter interval [1:1:85], and searched the hyper-parameter $\xi$ from a preset parameter interval [0.001:0.0002:0.01 0.01:0.002:0.1 0.1:0.02:1 1:0.2:10 10:2:100 100:20:1000].

\begin{figure*}[htbp]
    \centering
    \begin{overpic}[width=\textwidth]{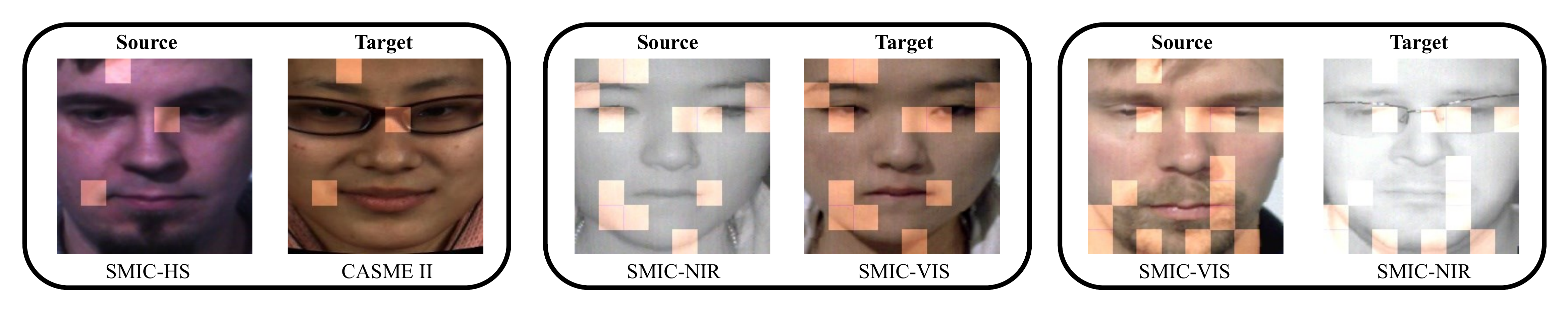}
    \end{overpic}
    \caption{
    The salient facial regions selected by our proposed TGSR method in three cross-database micro-expression recognition tasks.
    The sub-matrix $\hat{\varMatrix{C}}_i$ corresponding to the salient regions is the matrix filled with scalar one.
    }
    \label{fig:visualization}
\end{figure*}

\subsection{Results and Analysis}
%
\subsubsection{Overall results}
We bold-lighted the best result of each experiment in \tabref{tab:exp1} and \tabref{tab:exp2}.
We observed that our proposed TGSR outperforms those state-of-the-art methods beyond half experiments and achieved the best performance in 7 of total 12 CDMER experiments.
It indicates that our proposed TGSR has the ability to cope with the CDMER problem effectively.
We also reported the hyper-parameters to achieve this performance.
In the TYPE-I experiments, from Exp.1 to Exp.6, the best M-F1 is achieved at the hyper-parameters ($\kappa$,$\xi$) value of (85, 0.0022), (46, 0.0036), (14, 4000), (85, 0.0044), (12, 44), (12, 280), respectively.
In the TYPE-II experiments, from Exp.7 to Exp.12, the best M-F1 is achieved at the hyper-parameters ($\kappa$,$\xi$) value of (62, 0.0012), (28, 0.0980), (85, 0.0030), (85, 0.0280), (85, 0.0016), (75, 0.0220), respectively.
The best performance tends to select the few but discriminative regions rather than all facial regions.
And our proposed TGSR also displays competitive performance on those experiments that do not work best.
It also indicates the effectiveness of our salient facial region selection strategy.

\begin{figure}[htbp]
    \centering
    \begin{overpic}[width=\columnwidth]{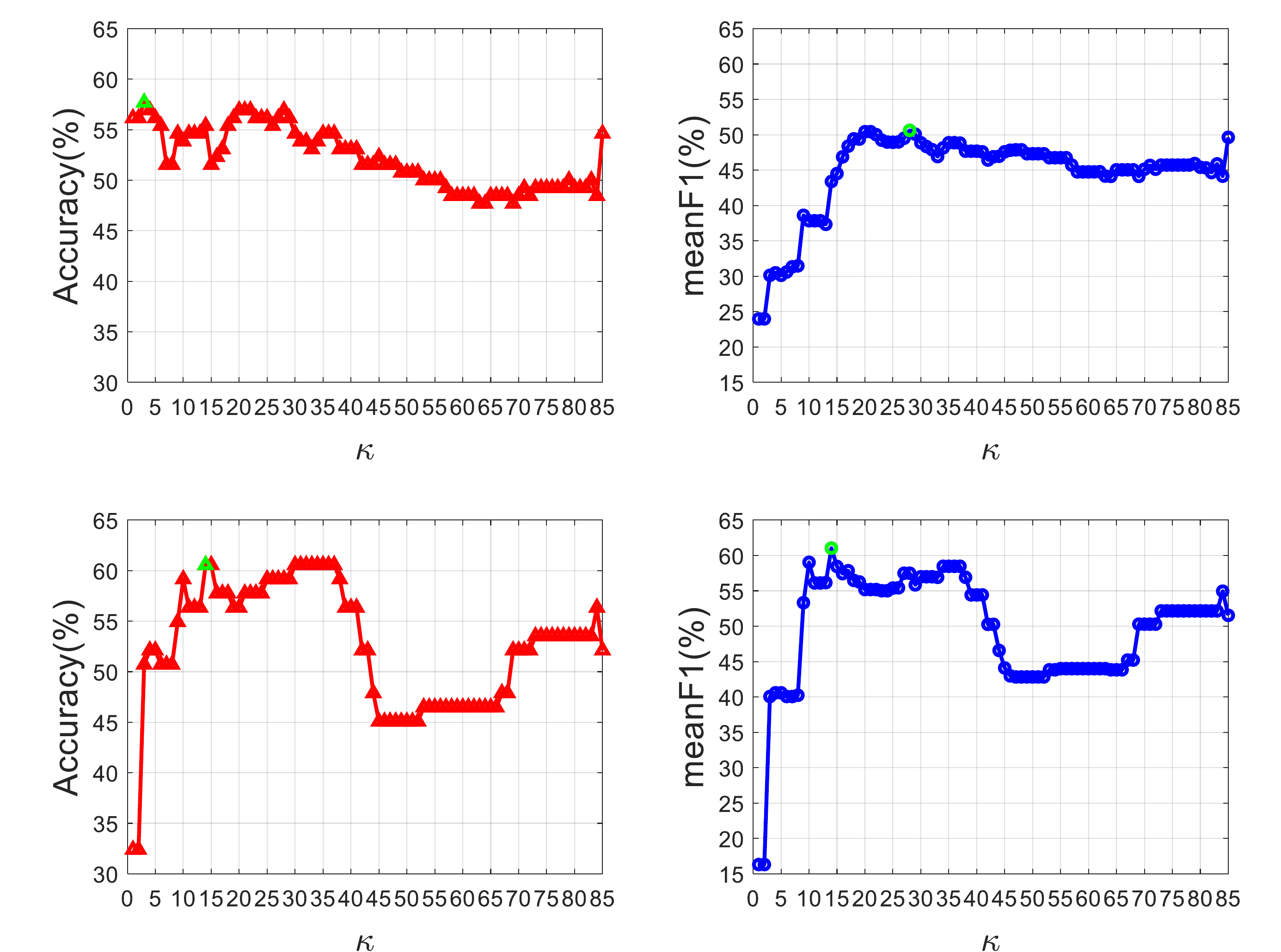}
        \put(0,55){(a)}
        \put(0,20){(b)}
    \end{overpic}
    \caption{
    The performance curve of our proposed TGSR method under different hyper-parameter $\kappa$, i.e., the salient facial region number.
    (a) shows the experimental results of Exp.8(H$\rightarrow$C) and (b) shows the experimental results of Exp.4(H$\rightarrow$N).
    }
    \label{fig:hyperpara_k}
\end{figure}

\begin{figure}[htbp]
    \centering
    \begin{overpic}[width=0.9\columnwidth]{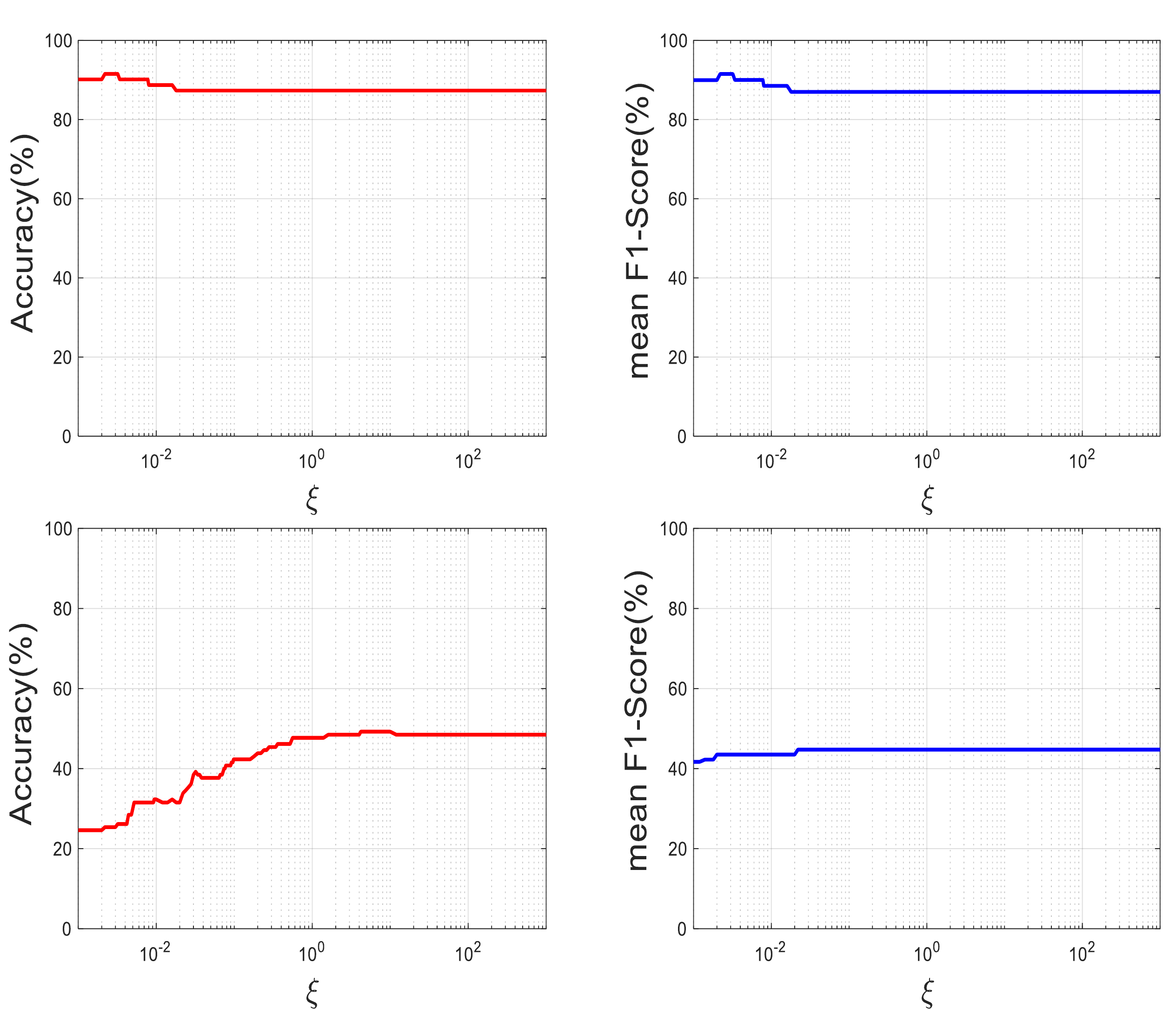}
        \put(-4,65){(a)} 
        \put(-4,23){(b)} 
    \end{overpic}
    \caption{
    The performance curve of our proposed TGSR method under different hyper-parameter $\xi$, i.e., weighting hyper-parameter of the MMD term.
    (a) shows the experimental results of Exp.1(H$\rightarrow$V) and (b) shows the experimental results of Exp.12(N$\rightarrow$C).}
    \label{fig:hyperpara_xi}
\end{figure}

\subsubsection{Difference analysis}
Two apparent performance characteristics involving database difference can be found in \tabref{tab:exp1} and \tabref{tab:exp2}.

%
Firstly, we observe that the results of TYPE-I experiments are generally better than TYPE-II experiments.
We believe the experimental setup itself caused it.
The TYPE-I experiments selected two subsets of SMIC database with different imaging modalities and TYPE-II experiments used Selected CASME II and one subset of SMIC database, as the source and target databases respectively.
It is clear that the database differences of TYPE-I experiments are more significant than TPYE-II experiments.

Secondly, we find that a noticeable performance gap existed in those experiments exchanging the source and target databases: all performance on Exp.1(H$\rightarrow$V) are generally better than those on Exp.2(V$\rightarrow $H); all performance on Exp.3(H$\rightarrow$N) are generally better than those on Exp.4(N$\rightarrow$H); all performance on Exp.11(C$\rightarrow$N) are generally better than those on Exp.12(N$\rightarrow$C).
Exp.1(H$\rightarrow$V) used the a high-speed camera captured image sequences from the SMIC-HS subset as the source database and the general visual camera captured image sequence from the SMIC-VIS subset as the target database, which is exchanged in Exp.2(V$\rightarrow$H).
Exp.3(H$\rightarrow$N) used colored image sequences captured by high-speed camera from SMIC-HS subset as the source database and the uncolored near-infrared image sequence from SMIC-NIR subset as the target database, which is exchanged in Exp.4(N$\rightarrow$H).
Exp.11(C$\rightarrow$N) used colored image sequences from Selected CASME II database as the source database and uncolored image sequences from SMIC-NIR subset as the target database, which is exchanged in Exp.12(N$\rightarrow$C).
%
We believe the reason why the performances on Exp.1(H$\rightarrow$V) are generally better than Exp.2(V$\rightarrow$H) and the performances on Exp.3(H$\rightarrow$N) are generally better than Exp.4(N$\rightarrow$H) is that high-speed camera can capture more subtle facial movements.
And the reason why the performances on Exp.11(C$\rightarrow$N) are generally better than Exp.12(N$\rightarrow$C) is that the Selected CASME II database retains but the SMIC-NIR subset discards color information such crucial cue for understanding human facial expressions\cite{benitez2018Facial}.
In addition, we observe that although Exp.3(N$\rightarrow$H)-Exp.4(N$\rightarrow$H) pair and Exp.11(C$\rightarrow$N)-Exp.12(N$\rightarrow$C) pair both have a color difference between the source and target databases, the model performance gap between Exp.11(C$\rightarrow$N)-Exp.12(N$\rightarrow$C) pair is more significant than Exp.3(N$\rightarrow$H)-Exp.4(N$\rightarrow$H) pair.
%
It may be due to other database differences other than the color.

\subsubsection{Hyper-parameter Discussion}
Two hyper-parameters are involved in solving the optimal regression matrix $\hat{\varMatrix{C}}$ of proposed our proposed TGSR, i.e., the salient facial region number $\kappa$ and the MMD weighting hyper-parameter $\xi$.
The setting of these hyper-parameters affects model performance, thus we conducted two experiments to investigate the model sensitiveness to hyper-parameters $\kappa$ and $\xi$.

\textbf{The number of salient facial regions.}
In the first experiment, we fixed hyper-parameter $\xi$ and varied hyper-parameter $\kappa$ from $1$ to $K=85$, then recorded corresponding M-F1 and ACC metrics, to explore model the sensitiveness to hyper-parameter $\kappa$.
We selected Exp.4(H$\rightarrow$N) and Exp.8(H$\rightarrow$C) as the typical of TYPE-I and TYPE-II CDMER experiments respectively, and presented their performance curve as \figref{fig:hyperpara_k} shown.
We can see that the M-F1 and ACC performance of our proposed TGSR model increases with hyper-parameter $\kappa$ increases firstly, and reaches its peak at a low $\kappa$ value, then decreases with hyper-parameter $\kappa$ increases.
It means that the salient facial regions for CDMER are exiguous.
And it also verifies the effectiveness of salient facial region selection.

\textbf{The weighting hyper-parameter of MMD term.}
In the second experiment, we fixed hyper-parameter $\kappa$ and varied hyper-parameter $\xi$ from $10^{-3}$ to $10^3$, then recorded corresponding M-F1 and ACC metrics, to explore the model sensitiveness to hyper-parameter $\xi$.
We selected Exp.1(H$\rightarrow$V) and Exp.12(N$\rightarrow$C) as the typical of TYPE-I and TYPE-II CDMER experiments respectively, and presented their performance curve as \figref{fig:hyperpara_xi} shown.
It is apparent that selecting an appropriate value of weighting hyper-parameter $\xi$ helps our proposed TGSR yield better performance.
And the MMD term can effectively and stably improve model performance across a wide range of hyper-parameter $\xi$.
%
%

\subsubsection{Visualization}
We also selected Exp.5(V$\rightarrow$N), Exp.6(N$\rightarrow$V), and Exp.8(H$\rightarrow$C) as the typical to visualize the learned salient facial regions for CDMER.
From \figref{fig:visualization}, we observed that the selected facial regions are consistent with the AU definition of micro-expressions; thus, we can believe that our proposed TGSR achieved a competitive performance by learning an explicable feature.


%


\section{Conclusion}
This paper proposes a novel Transfer Group Sparse Regression to select the salient facial regions to better cope with the Cross-Database Micro-Expression Recognition (CDMER) problem. 
In our proposed TGSR, salient facial region selection is achieved by the group feature selection from the raw face feature.
This operation enables our proposed TGSR to 1) optimize the measurement and alleviate the difference between the source and target databases and 2) highlight the valid facial regions to enhance extracted features.
In addition, this operation can also reduce computational costs while improving performance.
Experiments and visualizations show that our proposed TGSR learns the discriminative facial regions and outperforms most state-of-the-art subspace-learning-based domain-adaptive methods for CDMER.

\section*{Acknowledgment}
This work was supported in part by the National Natural Science Foundation of China (NSFC) under the Grants U2003207, 61921004, and 61902064, in part by the Fundamental Research Funds for the Central Universities under Grant 2242022k30036, and in part by the Zhishan Young Scholarship of Southeast University.






%

{\balance
\bibliographystyle{IEEEtran}
\bibliography{IEEEabrv,MyAbrv,egbib}
}

\end{document}